\documentclass{article}
\usepackage{textgreek}

\usepackage[dblblindworkshop, final]{neurips_2025}

\usepackage[utf8]{inputenc} 
\usepackage[T1]{fontenc}    
\usepackage{hyperref}       
\usepackage{url}            
\usepackage{booktabs}       
\usepackage{amsfonts}       
\usepackage{nicefrac}       
\usepackage{microtype}      
\usepackage{xcolor}         
\usepackage{siunitx}
\usepackage{graphicx}
\usepackage{footnote}
\makesavenoteenv{table}

\title{DynaPPI: A large-scale dynamic protein dataset for AI-driven advances in protein interactomics}
\workshoptitle{AI4Science}

\author{
    Jiabao Wei\textsuperscript{1}\thanks{Jiabao Wei is the project leader.}, Zilong Geng\textsuperscript{3}, Yuze Wang\textsuperscript{3}, Jianjun Li\textsuperscript{2}, \\\textbf{Ning Ding\textsuperscript{4,5}, Bowen Zhou\textsuperscript{4,5}, Bing Zhang\textsuperscript{3}, Zhiyuan Ma\textsuperscript{2}\thanks{Corresponding author.}}
    \vspace{1mm} \\
    \textsuperscript{1}BIT, 
    \textsuperscript{2}HUST, 
    \textsuperscript{3}SJTU, 
    \textsuperscript{4}Tsinghua University, 
    \textsuperscript{5}Shanghai AI Laboratory
}

\begin{document}

\maketitle

\begin{abstract}
Diffusion models have been widely explored in protein backbone generation due to their powerful generation capabilities.
However, in today's AI-driven biological research, predicting the structure of unknown multi-chain protein aggregates (called “\textit{complexes}” in biology) remains an unsolved challenge.
This is because existing static or dynamic protein datasets focus solely on static snapshots or single-entity trajectories, neglecting the dynamic process of multiple monomers forming complexes.
To alleviate this dilemma, we present \textbf{DynaPPI}, a dynamic protein dataset comprising molecular dynamics (MD) trajectories of protein complex formation from dissociated chains to the bound state, as a pivotal resource to bridge the gap between static structural biology and the inherently temporal nature of dynamic molecular interactions.
Benefiting from this dataset, diffusion models can explicitly learn the dynamic binding trajectories of known complexes and accurately predict the structures of unknown complexes based on their diverse generative properties, thereby further catalyzing AI-driven structural biology and protein interactomics.

\end{abstract}

\section{AI Task Definition}
The primary AI task associated with our dataset is defined as a \textbf{\textit{conditional generative prediction task}}, wherein diffusion models are trained to predict and generate realistic dynamic binding trajectories conditioned on the initial dissociated state and contextual biophysical parameters.
This task amalgamates elements of generation---generating novel trajectories that explore diverse binding modes---and prediction---predicting physically plausible outcomes based on empirical or simulated groundtruths.
Specifically, given the sequences, structures, or representations of these unbound protein chains, along with environmental conditions (\textit{e.g.}, temperature, ionic strength, or pH), the model predicts the time-resolved sequence of intermediate states and then generates the final composite structure.

\section{Dataset Rationale}
\noindent \textbf{Bottleneck:} Our dataset addresses a critical unmet need in AI-driven biology, where the transition from dissociated entities to the bound state remains poorly represented in existing resources.
Current datasets, such as PDB~\cite{berman2000pdb} or Dynamic PDB~\cite{liu2024dynamic}, predominantly focus on static snapshots or single-entity trajectories, capturing equilibrium structures but failing to elucidate the kinetic pathways, intermediate states, and non-additive emergent behaviors inherent to complex formation.
Our DynaPPI dataset is envisioned as a comprehensive repository of time-resolved recombination trajectories, encompassing not only multi-chain protein assemblies but also protein-ligand, protein-nucleic acid, and intra-molecular domain rearrangements.
By integrating MD simulations with experimental validations, this dataset promotes diffusion models to generate unknown-complex structures, ultimately enabling breakthroughs in fields such as drug design, synthetic biology, and systems pharmacology.

\begin{figure*}[t]
\centering
\includegraphics[width=\linewidth]{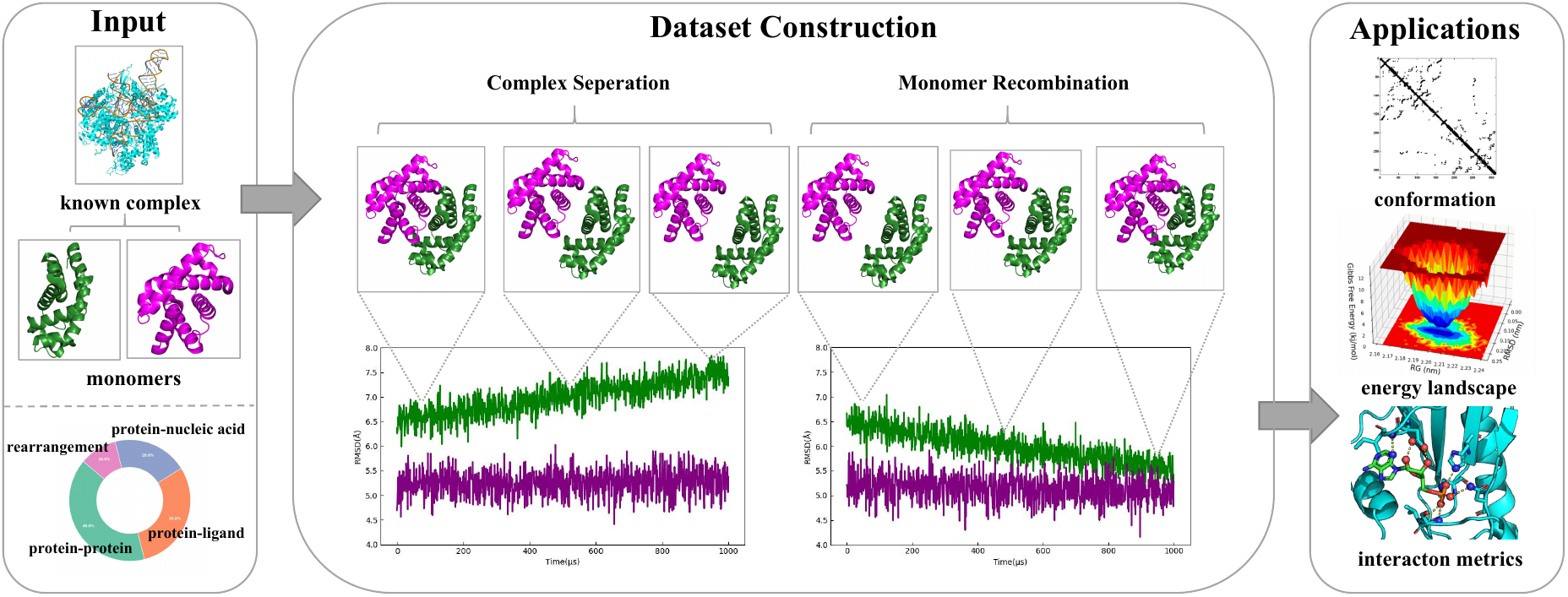}
\vspace{-10pt}
\caption{Overview of our dataset---input, construction, and applications. We disassemble known complexes from the PDB as references and record their MD recombination trajectories. Meanwhile, our dataset covers multiple categories of protein-related complexes, including protein-protein, protein-ligand, protein-nucleic acid, and intra-molecular rearrangements. We first explore the feasibility and effectiveness of the protein-protein split and provide a complete construction pipeline in Section~\ref{Multi-Chain Protein}.}
\label{overview}
\vspace{-15pt}
\end{figure*}

\noindent \textbf{Data types:} The dataset encompasses multi-modal data types, including time-series trajectories of atomic coordinates and physical properties, structural snapshots of clustered intermediate states, and auxiliary representations such as contact maps and free energy profiles, all formatted for efficient storage and AI compatibility (\textit{e.g.}, DCD$/$NetCDF for trajectories and HDF5 for properties).

\noindent \textbf{Scale:} The dataset targets about $10,000$ complexes initially, with $5\sim20$ replicas each spanning from $10$ ns to $1$ ms, yielding totally $10^8\sim10^{11}$ frames, covering diverse biomolecular interactions.
Moreover, the dataset ensures broad coverage across categories like protein-protein ($40\%$), protein-ligand ($30\%$), protein-nucleic acid ($20\%$), and intra-molecular rearrangements ($10\%$), with low sequence redundancy ($<30\%$) and storage in the $10\sim100$ TB range for scalable open access.

\noindent \textbf{Resolution:} Temporal resolution is maintained at $1\sim10$ ps per frame to capture rapid conformational changes while controlling data volume, complemented by all-atom spatial precision (sub-Å accuracy in coordinates and forces) for detailed modeling of non-additive interactions, with optional coarser variants (\textit{e.g.}, Cα-only backbones at $100$ ps) for computational efficiency.

\noindent \textbf{Labels and metadata needed:} Essential labels include state indicators (\textit{e.g.}, unbound$/$bound classifications), kinetic parameters, and thermodynamic metrics, supported by metadata such as source identifiers, environmental conditions, validation benchmarks, and AI-ready annotations like train$/$test splits and pre-computed embeddings to achieve robust, physics-informed model training.

\section{Acceleration Potential}
The far-reaching implications of our proposed dataset will extend to transformative applications in biomedicine and materials science.
In drug discovery, it could rapidly simulate ligand-binding pathways, effectively identify transient pockets for allosteric inhibitors and accurately predict off-target interactions in polypharmacology.
For synthetic biology, generative predictions could guide the design of self-assembling nanostructures or engineered enzymes with tunable affinities.
Moreover, by incorporating multi-modal data (\textit{e.g.}, integrating cryo-EM snapshots or fluorescence resonance energy transfer kinetics), the dataset further fosters interdisciplinary advancements, such as AI-accelerated virtual screening for pandemic preparedness or the rational engineering of biomolecular machines.


\section{Scalability}
The dataset exhibits strong scalability via leveraging parallelized MD simulations across distributed computing frameworks or using neural network potentials to approximate force fields.
Key enablers include modular curation (automated PDB filtering and preprocessing), batch processing for replicas, and \textit{\textbf{incremental growth by incorporating diverse biomolecular types}}.
At scale, it could encompass millions of trajectories, akin to the expanded version of Dynamic PDB, supporting broad AI training.
\section*{Acknowledgement}
This work is supported by the National Natural Science Foundation of China (No. 62406161), China Postdoctoral Science Foundation (No. 2023M741950), and the Postdoctoral Fellowship Program of CPSF (No. GZB20230347).

\bibliographystyle{plain}
\bibliography{refs}

\newpage
\appendix

\section{Dynamic Recombination Trajectory Dataset for Protein Complexes}
\label{Multi-Chain Protein}
Inspired by Dynamic PDB~\cite{liu2024dynamic}, we propose the construction of a \textbf{\textit{Dynamic Recombination Trajectory Dataset for Protein Complexes}} to capture the temporal evolution of protein complexes from separated chains to the bound state.
This innovative dataset extends the pipeline of Dynamic PDB~\cite{liu2024dynamic} to multi-chain systems while addressing the non-additivity of the binding process.
Specifically, the dataset focuses on the preliminary feasibility of about $100$ to $1,000$ complexes, simulated utilizing all-atom molecular dynamics (MD) simulation with OpenMM~\cite{eastman2017openmm7} for GPU acceleration and parallel optimization.
Moreover, production runs are set from $10$ ns to $1$ ms per replica, inconsistent with Dynamic PDB's~\cite{liu2024dynamic} $1$ μs simulations, since binding events are typically uncertain, but monitored in real time and dynamically adjustable.
Multiple replicas ensure stochastic sampling of binding paths.

In this section, we introduce the proposed dataset, detailing the preparation process, the molecular dynamics simulation, and the analysis of dynamic behaviors, respectively.

\subsection{Source Selection and Data Curation}
The original structures for the Dynamic Recombination Trajectory Dataset for Protein Complexes are sourced from the Protein Data Bank (PDB)~\cite{berman2000pdb}, focusing on multi-chain entries with an experimental resolution of no greater than $2.5\text{Å}$ determined via X-ray diffraction to ensure high structural reliability.
And in order to enhance annotation and diversity, we supplement PDB data~\cite{berman2000pdb} with specialized interaction databases, including PDBe-KB~\cite{varadi2020pdbe} for biological assemblies, BioLiP~\cite{yang2012biolip} for protein-protein interfaces (adapted from ligand contexts), and the Protein-Protein Docking Benchmark (version $5.0$)~\cite{vreven2015updates}, which provides over $230$ non-redundant complexes with validated binding modes.

Selection criteria prioritize systems amenable to molecular dynamics (MD) while promoting diversity: To maintain computational tractability, protein complexes are restricted to $2\sim4$ chains, with individual chain length $\leq300$ residues and a total system size $\leq50,000$ atoms (including solvent).
The dataset composition consists of $40\%$ homo-oligomers (\textit{e.g.}, symmetrical dimers such as HIV-1 protease), $40\%$ hetero-complexes (\textit{e.g.}, barnase-barstar complex for high-affinity binding), and $20\%$ transient interactions (\textit{e.g.}, ubiquitin-conjugating enzyme pairs).
This balance ensures representation across interface types, such as hydrophobic cores and electrostatic complementarity, as well as functional classes like enzyme-inhibitor and signaling interactions.
Furthermore, we exclude transmembrane complexes, large assemblies ($>4$ chains), and covalently linked systems to minimize simulation artifacts, focusing instead on non-covalent, soluble proteins.

For each selected complex, individual chains are first extracted from the bound PDB structure~\cite{berman2000pdb} and isolated to simulate the unbound state (more details in Section \ref{separation}).
Then these separate chains are translated $5\sim20$ nm apart with randomized orientations per replica, employing some toolkits like MDAnalysis~\cite{michaud2011mdanalysis} or VMD~\cite{humphrey1996vmd} for coordinate manipulation.
This configuration can effectively mimic diffusive encounters, achieving clear observation of the approach, collision, and binding phases.

The pilot phase targets $500$ complexes (\textit{e.g.}, $250$ dimers and $250$ trimers), with the final dataset expanding to about $5,000$ entries.
To avoid redundancy, complexes are further clustered by sequence similarity ($<30\%$) using MMseqs2~\cite{steinegger2017mmseqs2}.
Additionally, metadata, including PDB IDs, chain counts, and literature-derived affinities, are tracked in a SQLite database.
And the original structures can be downloaded via the PDB API or the official website\footnote{https://www.rcsb.org/}, with Biopython~\cite{cock2009biopython} for parsing.

\subsection{Preprocessing of Protein Data}
Prior to simulations, the original structures must undergo rigorous preprocessing to address PDB limitations such as missing residues and non-standard elements.
The specific gap \textbf{\textit{repair}} scheme is as follows: for segments $\leq5$ residues, we utilize MODELLER~\cite{vsali1993modeller} for homology-based loop modeling, combined with DOPE scoring~\cite{shen2006dope} for energetic evaluation; for larger gaps ($>5$ residues) or entire loops, AlphaFold-Multimer~\cite{evans2021AlphaFold-Multimer} can provide context-aware predictions, accessible via ColabFold.

\textbf{\textit{Cleaning}} involves: \textbf{\textit{1)}} removing all heteroatoms from the protein structures, including water molecules, ligands, and metal ions, to focus on the dynamic behavior between proteins; \textit{\textbf{2)}} mapping non-standard residues (\textit{e.g.}, selenomethionine to methionine) with Open Babel~\cite{o2011open}; and \textit{\textbf{3)}} protonating at physiological pH $7$ using PROPKA3~\cite{olsson2011propka3} or PDB2PQR~\cite{dolinsky2004pdb2pqr} for accurate charge assignment.
Moreover, explicit hydrogen atoms are added via MODELLER~\cite{vsali1993modeller} to achieve all-atom completeness.

We further \textbf{\textit{validate}} the preprocessing statistics, including completion rates (\textit{e.g.}, percentage relying on AlphaFold \textit{vs}. MODELLER) and gap size distributions, aiming for a rejection rate $<10\%$ due to irreparable issues.
After cleaning, isolated chains may undergo a brief energy minimization to relax strains.
And initial configurations for the separated state incorporate random rotations and translations via PyMOL scripts~\cite{delano2002pymol}, with clash detection leveraging MolProbity~\cite{chen2010molprobity} to confirm no structural overlaps.
Structural integrity also need to be validated with PROCHECK~\cite{laskowski1993procheck}, ensuring Ramachandran plot outliers $<1\%$; structures exceeding $5\%$ issues are discarded.

\begin{table}[t]
  \caption{Attributes of the proposed dataset.}
  \vspace{-3pt}
  \label{attributes}
  \centering
  \renewcommand\arraystretch{1.2}
  \scalebox{0.78}{\begin{tabular}{c|c|c|c|c}
    \toprule
    \textbf{Name} & \textbf{Data Type} & \textbf{Shape} & \textbf{Description} & \textbf{Unit} \\
    \midrule
    \multicolumn{5}{c}{\textit{Structural Information}} \\
    \midrule
    Replica ID & int8 & ($1$) & Identifier of Replica & - \\
    Position & float32 & ($N_{frames}$, $N_{atoms}$, $3$) & Trajectory Coordinates & $\text{Å}$ \\
    Distance & float32 & ($N_{frames}$, $N_{chains}*(N_{chains}-1)/2)$ & Inter-Chain Distances\footnote{The distance denotes pairwise center-of-mass distances between chains.} & $\text{Å}$ \\
    Contact Map & bool & ($N_{frames}$, $N_{residues}$, $N_{residues}$) & Binary Contact Maps\footnote{Binary contact maps ($1$ if distance $\leq4.5\text{Å}$, $0$ otherwise) for inter- and intra-chain interactions.} & - \\
    \midrule
    \multicolumn{5}{c}{\textit{Dynamic and Physical Property}} \\
    \midrule
    Velocity & float32 & ($N_{frames}$, $N_{atoms}$, $3$) & Trajectory Velocities & $\text{Å}/$ps \\
    Force & float32 & ($N_{frames}$, $N_{atoms}$, $3$) & Trajectory Forces & $\text{kcal}/\text{mol}\cdot\text{Å}$ \\
    Potential Energy & float32 & ($N_{frames}$) & System Potential Energy & $\text{kJ}/\text{mol}$ \\
    Kinetic Energy & float32 & ($N_{frames}$) & System Kinetic Energy & $\text{kJ}/\text{mol}$ \\
    Total Energy & float32 & ($N_{frames}$) & System Total Energy & $\text{kJ}/\text{mol}$ \\
    Temperature & float32 & ($N_{frames}$) & System Temperature & $\text{K}$ \\
    Pressure & float32 & ($N_{frames}$) & System pressure & $\text{bar}$ \\
    Box Volume & float32 & ($N_{frames}$, $3$) & System Volume Forces & $\text{nm}^3$ \\
    Density & float32 & ($N_{frames}$) & System Density & $\text{g}/\text{ml}$ \\
    \midrule
    \multicolumn{5}{c}{\textit{Binding-Specific Metric}} \\
    \midrule
    Interaction Energy & float32 & ($N_{frames}$, $N_{chains}*(N_{chains}-1)/2)$ & Inter-Chain Energy\footnote{This energy denotes decomposed inter-chain interaction energy via MMPBSA.} & $\text{kJ}/\text{mol}$ \\
    Hydrogen Bond & int16 & ($N_{frames}$) & Number of Hydrogen Bonds & - \\
    Salt Bridge & int16 & ($N_{frames}$) & Number of Salt Bridges & - \\
    Binding Status & bool & ($N_{frames}$) & Binding Status Flag & - \\
    $\Delta$SASA\footnote{SASA denotes solvent-accessible surface area.} & float32 & ($N_{frames}$, $N_{chains}$) & SASA Changes & $\text{Å}^2$ \\
    Final Status & bool & ($1$) & Status for Prolongation\footnote{This status indicates whether the simulation has extended beyond initial duration.} & - \\
    \bottomrule
  \end{tabular}}
  \vspace{-6pt}
\end{table}

\subsection{MD Simulation Setup and Execution}
The simulation environment is designed to mimic physiological conditions, extending Dynamic PDB's setup~\cite{liu2024dynamic} for multi-chain mobility.
Specifically, all-atom molecular dynamics simulations are conducted using OpenMM~\cite{eastman2017openmm7} in conjunction with the Amber-ff14SB force field~\cite{maier2015ff14sb}, which effectively governs protein interactions and further enhances the accuracy of protein side chain and backbone parameters.
Meanwhile, isolated protein chains are solvated in a truncated octahedral or cubic periodic box with a padding thickness of $\geq1.5$ nm to prevent self-interaction artifacts during diffusion.
And this box is filled with TIP3P water molecules for hydration accuracy, and subsequently neutralized and salted with Na$^+$$/$Cl$^-$ ions at a concentration of $150\ mM$.

Equilibration begins with an energy minimization process via steepest descent followed by conjugate gradient, targeting a force tolerance of $2.39\ kcal/mol\cdot\text{Å}$ (up to $10,000$ steps) to resolve bad contacts.
The canonical ensemble (NVT) phase runs for $1$ ns at $300\ K$ using the \textit{LangevinMiddleIntegrator} (friction coefficient of $1.0$ ps$^{-1}$ and time step of $2$ fs with SHAKE constraints~\cite{ryckaert1977shake} for hydrogen atoms), initially restraining protein heavy atoms (force constant of $10\ kcal/mol\cdot\text{Å}^2$ with gradual release).
And the isothermal-isobaric ensemble (NPT) phase follows for $1$ ns at $1$ bar, employing the \textit{Monte Carlo Barostat} (updates every $100$ steps) with the same integrator to equilibrate density.

Production runs are set from $10$ ns to $1$ ms per replica, scaled based on complex size and anticipated binding timescales (\textit{e.g.}, fast binders like barnase-barstar might stabilize in $10$ ns).
The hydrogen mass redistribution in OpenMM~\cite{eastman2017openmm7} is implemented with a time step of $2$ fs to achieve numerical stability.
Moreover, to account for energy barriers during binding and capture non-additive effects such as induced fit, enhanced sampling techniques are innovatively integrated: umbrella sampling~\cite{torrie1977umbrella} applies bias potentials along the center-of-mass distance reaction coordinate, while metadynamics~\cite{laio2002metadynamics} (via the PLUMED plugin in OpenMM~\cite{eastman2017openmm7}) adds history-dependent biases.
For diffusion-limited phases, $5\sim20$ replicas per complex are seeded with different random seeds, with Temperature replica-exchange (T-REMD)~\cite{sugita1999t-remd} performed in the $300\sim350\ K$ range to facilitate barrier crossing.

Real-time monitoring detects binding (\textit{e.g.}, $>20$ inter-chain contacts) via custom scripts, allowing for early termination of converged runs to optimize computation.
The simulations are run on NVIDIA A100 GPUs with $80$ GB of memory and parallelized with MPI in OpenMM~\cite{eastman2017openmm7}.
During the pilot phase, the total GPU hours per machine are expected to be $50\sim200$.

\subsection{Data Recording and Physical Properties}
Data recording intervals balance resolution and storage: atomic coordinates are recorded every $10$ ps and physical properties are recorded every $1$ ps, yielding $1$K$\sim10$K frames per ns (approximately $10$ GB per complex).
Table~\ref{attributes} provides a detailed overview of the data attributes associated with each replica, containing structural information, dynamic and physical properties, and binding-specific metrics.
This structured format supports subsequent analyses and interpretations of the inter-chain binding dynamic
behaviors and properties of the complexes within the dataset.

Metadata identifies binding events (\textit{e.g.}, timestamps of stable contacts), replica IDs, and prolongation status for extended runs.
Trajectories are stored in DCD or NetCDF formats, and attributes are stored in HDF5 for efficient access.
Ultimately, all data is compressed with gzip.

\subsection{Post-Processing, Analysis, and Validation}
Post-simulation processing aligns trajectories to reference bound structures using MDAnalysis~\cite{michaud2011mdanalysis}, thereby correcting periodic boundary artifacts.
States (i.e., unbound, transient intermediates, and bound) are identified via time-lagged independent component analysis (tICA)~\cite{naritomi2011tica} and k-means clustering with PyEMMA~\cite{scherer2015pyemma}, enabling quantification of transition rates using Markov State Models.

The analysis emphasizes non-additivity: per-chain and complex-wide RMSD$/$RMSF track conformational changes; evolution of the radius of gyration illustrates compactness shifts; free energy landscapes are reconstructed via WHAM~\cite{kumar1992wHAM} on biased simulations; and interface remodeling is assessed through variations in residue contact frequency.

Further quantitative validation ensures the dataset's internal quality (\textit{e.g.}, energy drift $<1\%$, RMSF alignment with Dynamic PDB single-chain benchmarks), as well as external quality (\textit{e.g.}, bound-state RMSD $<3\text{Å}$ \textit{vs}. experimental PDBs; binding affinities from MMPBSA correlating $>0.7$ \textit{vs}. experimental values from literature; and association rates from Markov State Models \textit{vs}. experimental values from literature).
And subsets are benchmarked against CAPRI~\cite{janin2003capri} and Docking Benchmark $5.0$~\cite{vreven2015updates} to assess pose accuracy.
Additionally, unstable runs (\textit{e.g.}, DSSP~\cite{kabsch1983dssp} secondary structure loss $>20\%$) are filtered, with a target success rate of $>80\%$.

The compiled dataset is split into train$/$validation$/$test ($80/10/10$), including raw trajectories, processed states, and access APIs, and will be hosted on Zenodo or Dryad with a DOI.

\subsection{Preliminary Timetable and Rough Budget}
The implementation unfolds in phases: selection and preprocessing ($6\sim8$ weeks, emphasizing manual curation), MD simulations ($8\sim16$ weeks, via automated batching), and analysis$/$validation ($4\sim6$ weeks, iterative refinement).
The rough budget ranges from $9,000\ \$\sim15,000\ \$$ for cloud computing (\textit{e.g.}, AWS p3.2xlarge at $3\ \$$ per hour), with open tools minimizing additional expenses; academic resources (\textit{e.g.}, NSF$/$NIH clusters with SLURM) could further reduce this for scale-up.

\section{Initial Separation of Reference Complexes}
\label{separation}
The initial separation of protein chains from bound complex structures in the PDB~\cite{berman2000pdb} is critical for simulating realistic recombination dynamics.
However, direct extraction from bound PDB entries retains conformational biases induced by inter-chain interactions, such as interface remodeling or induced-fit changes, which deviate from the native unbound (apo) states.
This can underestimate energy barriers and non-additive effects during binding, leading to artificially accelerated or biased trajectories.
To mitigate this, we prioritize or approximate unbound conformations, ensuring starting structures more accurately reflect free-chain dynamics.
The procedure comprises three sequential steps: (1) chain extraction, (2) unbound state approximation, and (3) geometric separation.

\noindent \textbf{Step 1: Chain Extraction.}
First, load the bound PDB structure into a molecular analysis framework.
After that, parse and isolate individual chains by selecting atoms based on chain identifiers (\textit{e.g.}, ‘A’, ‘B’).
Then export each chain as a separate PDB file, preserving atomic coordinates, residues, and any existing hydrogen atoms. Finally, validate completeness by checking for gaps or clashes using stereochemical assessment tools, rejecting structures with $>5\%$ unresolved residues.

\noindent \textbf{Step 2: Unbound State Approximation.}
Approximate native unbound conformations to remove bound-state bias: First, query databases (\textit{e.g.}, PDBe-KB~\cite{varadi2020pdbe}, UniProt~\cite{uniprot2025uniprot}) for experimental unbound structures of each chain (sequence identity $>95\%$).
If available, align to the bound reference using TM-align structural superposition~\cite{zhang2005tm} and adopt as the starting model.
If unbound structures are unavailable, extract chains from the bound PDB and relax via short molecular dynamics (MD) simulations: Solvate each chain individually in a TIP3P water box with $150\ mM$ NaCl, minimize energy (tolerance $2.39\ kcal/mol\cdot\text{Å}$), equilibrate (NVT$/$NPT, $1$ ns each at $300\ K$), and run production for $5\sim10$ ns with Langevin integration ($2$ fs timestep).
Cluster resulting frames (\textit{e.g.}, using tICA~\cite{naritomi2011tica} and k-means) and select the dominant conformation (RMSD $>1\text{Å}$ from bound state).
Moreover, as an alternative or augmentation for incomplete chains, predict unbound models from sequences using AlphaFold3~\cite{abramson2024AF3} in monomer mode, aligning outputs to bound equivalents for coordinate consistency.

\noindent \textbf{Step 3: Geometric Separation.}
Following the previous step, recombine approximated unbound chains into a single structure and compute centers of mass for each chain.
After that, apply random rotations (Euler angles uniformly sampled from $0\sim360^\circ$) and translations (vectors yielding $5\sim20$ nm inter-chain separation, randomized per replica) to mimic diffusive encounters.
Then assign initial velocities from a Maxwell-Boltzmann distribution at $300\ K$.
Ultimately, validate final configurations for clashes (minimum distance >0.5 nm) and export as a merged PDB file for solvation.

In conclusion, this elaborately designed process of complex separation effectively enhances simulation fidelity by starting from unbound-like states, better capturing non-additive binding effects, at a modest computational cost ($1\sim2$ GPU-hours per chain for relaxation$/$prediction).


\end{document}